\documentclass[preprint,12pt,authoryear]{elsarticle}

\usepackage{comment}
\usepackage{amssymb}
\usepackage{amsmath}
\usepackage{multirow}
\usepackage{xcolor}
\usepackage{hyperref}
\usepackage{graphicx}
\usepackage{longtable}
\usepackage{lscape}
\journal{ }

\begin{document}

\begin{frontmatter}



\title{Experimenting with Multi-modal Information to Predict Success of Indian IPOs} %% Article title


\author[label1]{Sohom Ghosh} %% Author name
\ead{sohomg.cse.rs@jadavuruniversity.in}
\author[label2]{Arnab Maji}
\ead{arnabmaji09@gmail.com}
\author[label3]{N Harsha Vardhan}
\ead{nemani.v@research.iiit.ac.in}
\author[label1]{Sudip Kumar Naskar}
\ead{sudipkumar.naskar@jadavpuruniversity.in}

\affiliation[label1]{organization={Jadavpur University},%Department and Organization
            %addressline={}, 
            city={Kolkata},
            %postcode={}, 
            %state={},
            country={India}}
\affiliation[label2]{organization={Independent Researcher},%Department and Organization
            %addressline={}, 
            city={Kolkata},
            %postcode={}, 
            %state={},
            country={India}}
\affiliation[label3]{organization={International Institute of Information Technology},%Department and Organization
            %addressline={}, 
            city={Hyderabad},
            %postcode={}, 
            %state={},
            country={India}}            
%% Abstract
\begin{abstract}
 With consistent growth in Indian Economy, Initial Public Offerings (IPOs) have become a popular avenue for investment. With the modern technology simplifying investments, more investors are interested in making data driven decisions while subscribing for IPOs. In this paper, we describe a machine learning and natural language processing based approach for estimating if an IPO will be successful. We have extensively studied the impact of various facts mentioned in IPO filing prospectus, macroeconomic factors, market conditions, Grey Market Price, etc. on the success of an IPO. We created two new datasets relating to the IPOs of Indian companies. Finally, we investigated how information from multiple modalities  (texts, images, numbers, and categorical features) can be used for estimating the direction and underpricing with respect to opening, high and closing prices of stocks on the IPO listing day.
\end{abstract}

\begin{graphicalabstract}
\includegraphics{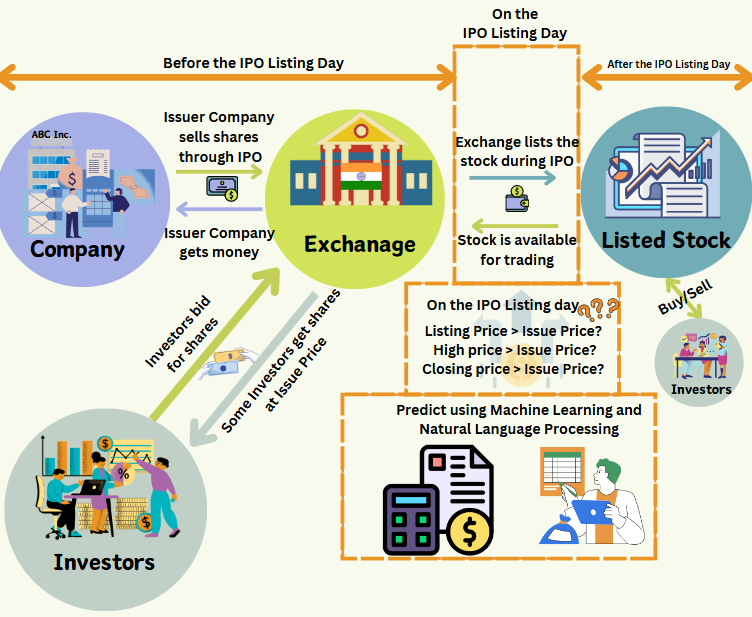}
\end{graphicalabstract}

\begin{highlights}
\item Curated two datasets on SMEs and Main Board IPOs listed in the Indian Stock Exchange
\item Used Machine Learning \& Natural Language Processing to predict success of Indian IPOs
\item Analysed effect of macroeconomic factors \& company financials on IPO success 
\item Studied DRHP and RHP data to predict IPO success \& GMP's trustworthiness
\item Proposed a multi-classifier system for estimating IPO stock prices on listing day

% \item Curated two multi-modal datasets consisting of details relating to Small and Medium Enterprise (SME), Main Board Initial Public Offerings (IPOs) of companies which were listed in the Indian Stock during the last decade.
% \item Used Multi-classier based Machine Learning systems and Natural Language Processing to predict if Initial Public Offerings IPOs in the Indian stock market would be successful
% \item Analysed impact of various macroeconomic factors, prevailing stock market's performance, and financials of a company towards the success of its IPO
% \item Investigated how information from Draft Red Herring Prospectus (DRHP) and Red Herring Prospectus (RHP) can be automatically mined for predicting success of these IPOs
% \item Studied the trust worthiness of Grey Market Premium (GMP) in estimating listing price of Indian IPOs
% \item Proposed a novel multi-classifier based decision-making system which takes performs information fusion over texts, numeric values, and categorical features for estimating the direction and underpricing with respect to opening, high and closing prices of stocks on the IPO listing day
\end{highlights}

%% Keywords

\end{frontmatter}

\section{Introduction}
Recently, with the growth in Indian economy, there is a huge surge in interest towards making investments in the stock market.\footnote{\href{https://www.businesstoday.in/markets/top-story/story/demat-accounts-at-all-time-high-cdsl-nse-gain-market-share-425233-2024-04-12}{https://www.businesstoday.in/markets/top-story/story/demat-accounts-at-all-time-high-cdsl-nse-gain-market-share-425233-2024-04-12} (accessed on 19\textsuperscript{th} August 2024)} This is due to factors like easing the investment process, allowing low ticket investments, providing liquidity, generating returns that help to hedge against inflation, etc. A private company transitions its private ownership to public trading though an Initial Public Offering (IPO). This allows the company to raise capital. Investors are allured towards subscribing for IPOs as it helps them to book profits quickly from the listing gains and provides them with access to early-stage companies. 

A company needs to submit Draft Red Herring Prospectus (DRHP) to the Securities and Exchange Board of India (SEBI). It contains  information regarding the company's fundamentals,  business, operations, financial performance, prospects, and legal issues. The DRHP is circulated to potential investors for initial evaluation and feedback. Later, it is finalized and presented as the Red Herring Prospectus (RHP). SEBI oversees and regulates the IPOs. This makes the process of filing an IPO transparent and instils confidence  among investors. 

The price at which the company's shares are first offered to the public during an IPO is called the Issue Price. There are primarily two types of IPOs in India: Fixed Price Issue, and Book Building Issue. In a fixed price IPO, the share price is established in advance and communicated to investors prior to the opening of the issue. This price is determined by the company in collaboration with the merchant bankers, taking into account various factors such as the company's valuation, assets, liabilities, risks, and growth potential. In a book building IPO, the company provides a price range (from a minimum to a maximum price) rather than a fixed price. Investors have the flexibility to place bids at any price within this range. The final Issue Price is established based on the bids collected after the issue period concludes. Some other types of IPOs are: Rights Issue and Follow-on Public Offer (FPO). Rights Issue allows existing shareholders to purchase new shares. FPO enables companies that are already listed on stock exchanges to raise additional capital. 

Based on size of the company, and issue sizes, IPOs can be categorized into Main Board (MB) IPO and Small and Medium Enterprises (SME) IPO. Compared to MB IPOs, SME IPOs have more relaxed eligibility criteria, allowing smaller companies to access public funding more easily. SME IPOs are primarily vetted by the respective stock exchanges (BSE SME or NSE Emerge), while MB IPOs require scrutiny and approval from SEBI, which includes a more comprehensive review of the prospectus. The differences between these two categories relate to paid-up capital, the minimum number of allottees, underwriting requirements, minimum application size, and market-making practices \footnote{\url{https://www.indiainfoline.com/knowledge-center/ipo/difference-between-mainboard-ipo-sme-ipo} (accessed on 23\textsuperscript{rd} August, 2024)}.

Overall, the Indian IPO landscape remains vibrant and dynamic. \footnote{\href{https://www.businesstoday.in/markets/ipo-corner/story/ipo-flood-rs-115-lakh-crore-worth-of-public-offers-likely-to-hit-markets-in-next-12-months-444559-2024-09-05}{https://www.businesstoday.in/markets/ipo-corner/story/ipo-flood-rs-115-lakh-crore-worth-of-public-offers-likely-to-hit-markets-in-next-12-months-444559-2024-09-05} (accessed on 6\textsuperscript{th} September, 2024)} Presently, India has been issuing the highest number of IPOs per year. \footnote{\href{https://www.livemint.com/market/stock-market-news/matter-of-great-pride-madhabi-puri-buch-says-india-ipo-issuances-rank-no-1-in-global-league-tables-11722777200397.html}{https://www.livemint.com/market/stock-market-news/matter-of-great-pride-madhabi-puri-buch-says-india-ipo-issuances-rank-no-1-in-global-league-tables-11722777200397.html} (accessed on 19\textsuperscript{th} August 2024)} However, factors like market volatility, over-subscription, pricing, influence of investor sentiment, and social media chatter may have adverse effects on the expected return, and the market premium. SME companies have begun to exploit the more lenient regulatory framework. Recently, several instances of fraud have emerged, leading SEBI to issue enforcement orders against some of these firms \footnote{\href{https://www.linkedin.com/pulse/jay-powell-says-let-party-continue-zerodha-dixtf/}{https://www.linkedin.com/pulse/jay-powell-says-let-party-continue-zerodha-dixtf/} (accessed on 27\textsuperscript{th} August, 2024)}. SEBI Chairperson, expressed her concerns regarding potential manipulation within the Small and Medium Enterprises (SME) segment. She noted that the market regulator has detected indications of such manipulation, highlighting feedback from the market that suggests misuse of SME listing provisions. \footnote{\url{https://finshots.in/archive/nse-cracks-down-on-shady-sme-ipos/} (accessed on 4\textsuperscript{th} September 2024)} These days, lots of retail investors are making speculative investments instead of relying on the fundamentals.\footnote{\url{https://www.moneycontrol.com/news/business/markets/raamdeo-agrawal-ola-electric-fundamentals-12801723.html} (accessed on 31\textsuperscript{st} August, 2024)}. Among the 10 largest first-day gainers in SME IPOs, nine have declined from their closing prices on day one. Additionally, 50\% of SME stocks experience a drop after their initial listing day gains.\footnote{\href{https://www.financialexpress.com/market/50-sme-stocksnbsp-stumble-after-listing-day-gains-3596400/}{https://www.financialexpress.com/market/50-sme-stocksnbsp-stumble-after-listing-day-gains-3596400/} (accessed on 31\textsuperscript{st} August, 2024)} Over half (54\%) of the IPO shares allocated to retail investors were sold within a week of the listing.\footnote{\url{https://www.moneycontrol.com/news/business/markets/sebi-study-retail-sold-ipo-12812542.html} (accessed on 3\textsuperscript{rd} September, 2024)} This, indicates that a large chunk of investors seeks listing gains. Many investors blindly trust the Grey Market Premium (GMP) \footnote{\url{https://www.chittorgarh.com/book-chapter/ipo-grey-market-gmp/28/} (accessed on 19\textsuperscript{th} August 2024)} for investing in an IPO. GMP refer to the difference between the Issue Price (the price at which shares are offered to the public) and the price at which the shares are traded in the unofficial and unregulated grey market. The Indian IPO market has witnessed significant growth in recent years, attracting speculative investors seeking to capitalize on the potential of emerging markets. These investors need to be educated. \footnote{\url{https://www.financialexpress.com/opinion/educate-retail-investors/3601919/} (accessed on 6\textsuperscript{th} September, 2024)} Thus, we need a framework for understanding the success of Indian IPOs through thorough examination of the various factors influencing their performance. 

Under-pricing in an IPO (Initial Public Offering) refers to the phenomenon where a company's shares are issued at a price lower than their actual market value, resulting in a significant increase in the share price on the first day of trading. The under-pricing percentage is calculated as: Under-pricing Cost = [(P\textsubscript{m} - P\textsubscript{0}) / P\textsubscript{0}] * 100, where P\textsubscript{m} is the closing price on the first day of trading and P\textsubscript{0} is the Issue Price. Most of the prior research work (\cite{bansal2012determinants},\cite{BASTI201515}, 
\cite{BAJO2017139}, \cite{quintana2018fuzzy}, \cite{ramesh2019revisiting}, \cite{sakharkar2019pricing}, \cite{BABA202013}) relating to IPO studied under-pricing of MB IPOs.

In this paper, we propose a Machine Learning (ML) and Natural Language Processing (NLP) based framework for determining if an IPO in the Indian market will be successful in the short term. We define success in terms of the difference between issue price and the opening price, high price, and closing price on the day of the IPO. We studied this separately for MB and SME IPOs. Furthermore, we investigate how much GMP of IPOs are trustable.

% \textcolor{red}{ADD MORE DETAILS REGARDING OUR METHODOLOGY}\\
% \textcolor{violet}{Narrate: What is the novelty? SEBI PDFs RAG, Helpful for traders as taking penultimate day's subscription rate not consideration}\\

%IPO Asia-Pacific trends: %\url{https://www.ey.com/en_gl/insights/ipo/trends}

\subsection*{Our Contributions}
\begin{itemize}
    \item We present two multi-modal datasets, one for Main Board IPOs, and the other for Small and Medium Enterprises (SME) IPOs. It consists of various features relating to the company going for IPOs, and other macroeconomic factors.
    \item We propose Machine Learning and NLP based decision system for predicting if an IPO will be successful
    \item We study the impact of various macroeconomic factors, prevailing stock market’s performance, and financials on the success of an IPO
    \item We extract important portions from documents like DHRP, RHP and used them as features for predicting the success of an IPO
    \item We investigate the relation between of GMP and success of an IPO 
    \item We evaluate the performance of Large Language Models (LLMs) like Gemini \cite{reid2024gemini} and Llama \cite{llama3modelcard} in predicting success of Indian IPOs
\end{itemize}

The reminder of this paper is organized as follows: related work
in presented in Section \ref{sec:related-work}. The problem has been described in
Section \ref{sec:problem-description}. The data preparation steps are mentioned in section \ref{sec:data-prep}. %and \ref{sec:methodology} respectively. 
Experiments are described in the Section \ref{sec:experiments}. Section \ref{sec:conclusion} concludes the paper.

\section{Related work}
\label{sec:related-work}
The landscape of Initial Public Offerings (IPOs) in India has been a focal point for researchers aiming to understand the various factors influencing their success. This literature review synthesizes findings from multiple studies, highlighting key themes such as underpricing, regulatory impacts, investor behaviour, and the overall performance of Indian IPOs.
For the last few decades, several researchers have studied the IPO market of various countries like India (\cite{bansal2012determinants}, \cite{ramesh2019revisiting}, \cite{sakharkar2019pricing}), the USA (\cite{BAJO2017139}, \cite{WANG2018102}, \cite{ly-2020-words-ipo}), China (\cite{CHI200571}), Turkey (\cite{BABA202013}, \cite{bateni2014study}), South Korea (\cite{Kim2010}), and Malaysia (\cite{wong2017initial}). Most of these studies were related to underpricing, its causes and effects (\cite{bansal2012determinants},\cite{BASTI201515}, 
\cite{BAJO2017139}, \cite{quintana2018fuzzy}, \cite{ramesh2019revisiting}, \cite{sakharkar2019pricing}, \cite{BABA202013}). 

\cite{seepani2023initial} presents a structural review of IPOs in India, covering the period from pre-liberalization to the present. This study reveals significant insights into the evolution of the IPO market. The research emphasizes that IPO volume and valuation are heavily influenced by regulatory changes \cite{yadav2019research}, economic growth \cite{ramesh2015performance} and global financial conditions. Notably, the fiscal year 2022-2023 witnessed a surge in IPO activity, driven by increased retail investor participation and a focus on technology and startups. This study underscores the importance of understanding India's unique socio-economic factors that shape the IPO landscape.

Most of the research works focus either to determine the short-run underpricing (\cite{ramit-ijcg-2019}, \cite{WANG2018102}, \cite{BAJO2017139}, \cite{iqbal-pricing-2018}, \cite{manu2020valuation}) or the long-run underperformance (\cite{sahoo-vikalpa-2010}). Traders are interested about short-run underpricing, while investors are interested about long-run performance. Factors such as a company's age \cite{bhatia2012examining}, pricing mechanism, retails subscription, market capitalization  \cite{bansal2012determinants}, size,  return on assets, financial leverage, and the reputation of its underwriters \cite{dewi2024factors} are crucial in determining the initial offering price. Additionally, political stability \cite{Mehmood2021determinants}, market conditions — including general economic sentiment \cite{ghosh2005underpricing} and industry trends \cite{india1initial} also have a substantial impact on IPO performance and pricing. Other factors such as group affiliation \cite{marisetty2006group}, regulatory environment \cite{yadav2019research}, and effective communication strategies with investors are recognized as critical in increasing the chances of IPO success. \footnote{\href{https://fastercapital.com/topics/the-role-of-communication-in-a-successful-ipo.html}{https://fastercapital.com/topics/the-role-of-communication-in-a-successful-ipo.html} (accessed on 1\textsuperscript{st} September 2024} A study \cite{sandhu2020effects} analysing the relationship between IPO offer price ranges and initial demand among investors indicates that lower-priced IPOs tend to attract less trading activity post-listing. This research highlights that institutional investors favour higher-priced offerings, while retail investors are more likely to subscribe to lower-priced IPOs. This behaviour contributes to the underpricing phenomenon, which is positively correlated with over subscription rates in the Qualified Institutional Buyers (QIB) category. Another empirical study \cite{ramit-ijcg-2019} explores the impact of board composition and promoter ownership on IPO underpricing. The findings suggest that reputable boards can mitigate information asymmetry, thereby reducing underpricing. Conversely, high promoter ownership is associated with increased underpricing, indicating potential conflicts of interest that may arise from insider control. Research \cite{sakharkar2019pricing} indicates that various factors influence the performance of Indian IPOs, including market conditions, pricing strategies, and the timing of offerings  \cite{sahoo-vikalpa-2010}, \cite{khatri2017factors}. A comprehensive analysis \cite{sakharkar2019pricing}  of 290 IPOs from 2007 to 2017 reveals significant underpricing, with an average raw return of 17.90\% in the short term, which declines sharply after nine months. This suggests that while IPOs may perform well initially, long-term investments may not yield favourable outcomes. Additionally, the study \cite{sakharkar2019pricing} highlights that IPOs issued after 2013 generally performed better, with shorter listing delays correlating with higher returns. The impact of offer price and size on performance is also notable, with mid-range offerings typically yielding better results \cite{ramesh2019revisiting}.
    
Research \cite{sahoo-vikalpa-2010} reveals that initial underpricing often leads to long-run under performance, with IPOs failing to maintain their initial momentum. While underpricing offers short-term gains, the long-term performance of Indian IPOs presents a more nuanced picture. \cite{mehta2016price}, \cite{mayur2014relationship}, and \cite{khan2021study} present a comparative analysis between the short term and long term performance.

The role of investor sentiment and macroeconomic conditions \cite{phadke2018impacts} in IPO performance are other critical areas of exploration. \cite{BAJO2017139} indicates that positive media sentiment can significantly influence retail investor perceptions and demand, thereby affecting first-day returns. This suggests that external perceptions play a crucial role in shaping IPO success and investor behaviour. A study \cite{NIKBAKHT2021100400} focusing on pre-IPO earnings management reveals that firms utilizing reputable investment banks are less likely to manipulate earnings, highlighting the importance of transparency in the IPO process. This research indicates that improved governance mechanisms can enhance investor confidence and potentially reduce underpricing. Several studies (\cite{sahoo-vikalpa-2010}, \cite{srinivasa2015aftermarket}) suggest that investors perceive underpricing as a signal of quality and future growth potential, leading to increased demand and subsequent price appreciation. However, the motivations for underpricing also include mitigating risk for issuers and attracting investors, with potential consequences for long-term performance \cite{mehta2016price}.

\cite{locke2009return} presents a comparative analysis of IPO returns between India and China illustrates distinct return patterns influenced by differing political and economic systems. While Indian IPOs tend to exhibit positive initial returns, the sustainability of these returns diminishes over time, emphasizing the need for investors to consider risk factors associated with IPO investments in emerging markets.

While most of the research papers deal primarily with numeric features, there are a few which uses text based features as well (\cite{BAJO2017139}, \cite{ly-2020-words-ipo}). \cite{BAJO2017139} studies sentiment of news articles, \cite{liew2016twitter} investigated effect of sentiments of tweets, while \cite{ly-2020-words-ipo} analyses IPOs’ prospectuses from the SEC database to estimate success of IPOs. Similarly, most of the research paper used regression models to predict the performance of the IPOs. Only a handful of research papers used advanced machine learning based algorithms like Support Vector Machines Random Forests \cite{BABA202013}, \cite{BASTI201515},  and fuzzy techniques \cite{quintana2018fuzzy}.

The literature on Indian IPOs presents a multifaceted view of the factors influencing their success. The prevalence of underpricing, the impact of regulatory changes, and the role of investor behaviour are critical themes that emerge from the research. However, there are certain research gaps which can be addressed. Firstly, conducting sector-specific analyses of IPO performance may reveal unique challenges and success factors that are not captured in broader studies. Secondly, most of the prior works defined under-pricing in terms of closing price of  the listing day. But, for benefiting short-term traders, it may be interesting to investigate how open price and highest price on listing day varied. Thirdly, the impact of grey market price and ratings of Indian IPOs by top analysts have not yet been thoroughly studied. Lastly, although Large Language Models (LLM) have shown remarkable performance in various financial tasks (\cite{xie2024finben}, \cite{xie2024openfinllms}), no one have ever used them to analyse DRHP and RHP of companies for predicting IPO related success. In this paper, we would like to address these research gaps.

\section{Problem statement}
\label{sec:problem-description}
We define the success of a company's IPO by comparing issue price with the opening price, high price, and closing price  on the listing day of the IPO.

\begin{itemize}

\item Opening price \\
 $\cdot$ Predict if the opening price on the listing day of the IPO will be greater than the issue price of the IPO. We refer to this as predicting the direction of opening price movement.\\
 $\cdot$ Predict underpricing with respect to opening price on the listing day of the IPO, i.e. (opening price - issue price)/(issue price)

\item High price\\
 $\cdot$ Predict if the highest price on the listing day of the IPO will be greater than the issue price of the IPO\\
 $\cdot$ Predict underpricing with respect to the highest price on the listing day of the IPO, i.e. (highest price - issue price)/(issue price)

\item  Closing price\\
$\cdot$ Predict if the closing price on the listing day of the IPO will be greater than the issue price of the IPO\\
 $\cdot$ Predict underpricing with respect to closing price on the listing day of the IPO, i.e. (closing price - issue price)/(issue price)

\end{itemize}

We conduct the separate experiments for SME and Main Board IPOs with the same objective of predicting the success of IPOs.

\section{Data preparation}
\label{sec:data-prep}
Firstly, we prepared a list of companies which went for Main Board IPO from 2009 to 2023. Similarly, we prepared another list of companies which went for SME IPO from 2017 to 2023.  We collected this data from chittorgarh.com \footnote{\url{https://www.chittorgarh.com/ipo/ipo_dashboard.asp} (accessed on 23\textsuperscript{rd} January, 2024)}. The date ranges were decided based on availability of the dataset. After removing all the instances where the IPO was withdrawn or no data was present, we were left with data for 418 Main Board, and 681 SME IPOs.

\begin{figure}[t]%% placement specifier
%% Use \includegraphics command to insert graphic files. Place graphics files in 
%% working directory.
\centering%% For centre alignment of image.
\includegraphics[width=\textwidth]{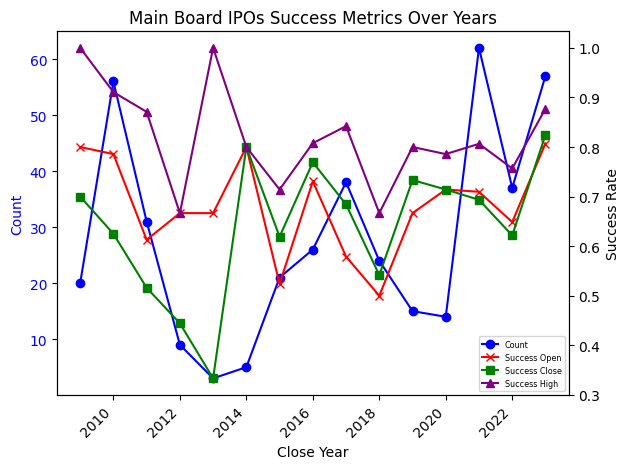}
%% Use \caption command for figure caption and label.
\caption{Success rates over the year for Main Board IPOs}
\label{fig:main-success}
\end{figure}

\begin{figure}[t]%% placement specifier
%% Use \includegraphics command to insert graphic files. Place graphics files in 
%% working directory.
\centering%% For centre alignment of image.
\includegraphics[width=\textwidth]{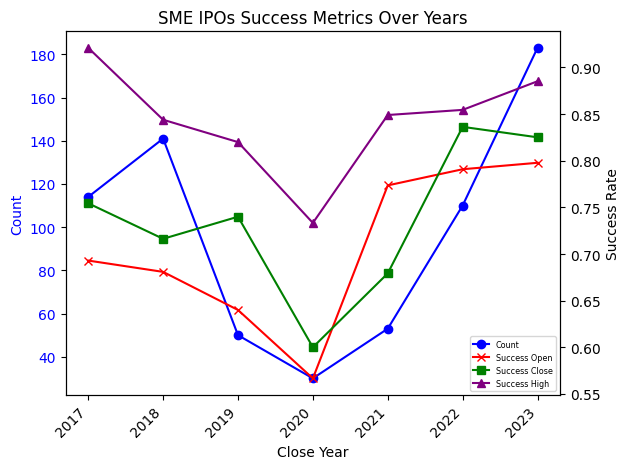}
%% Use \caption command for figure caption and label.
\caption{Success rates over the year for SME IPOs}
\label{fig:sme-success}
\end{figure}

For both SME and Main Board IPOs, to train the models, we used the data till 2022. The data for the year 2023 was used to evaluate the performances of the trained models. We present the train, test split in Table \ref{tab:data-dist}. Figures \ref{fig:main-success} and \ref{fig:sme-success} represent how the success rate of Main Board and SME IPOs respectively varied over the years.

\begin{figure}[t]
\centering
\includegraphics[width=\textwidth]{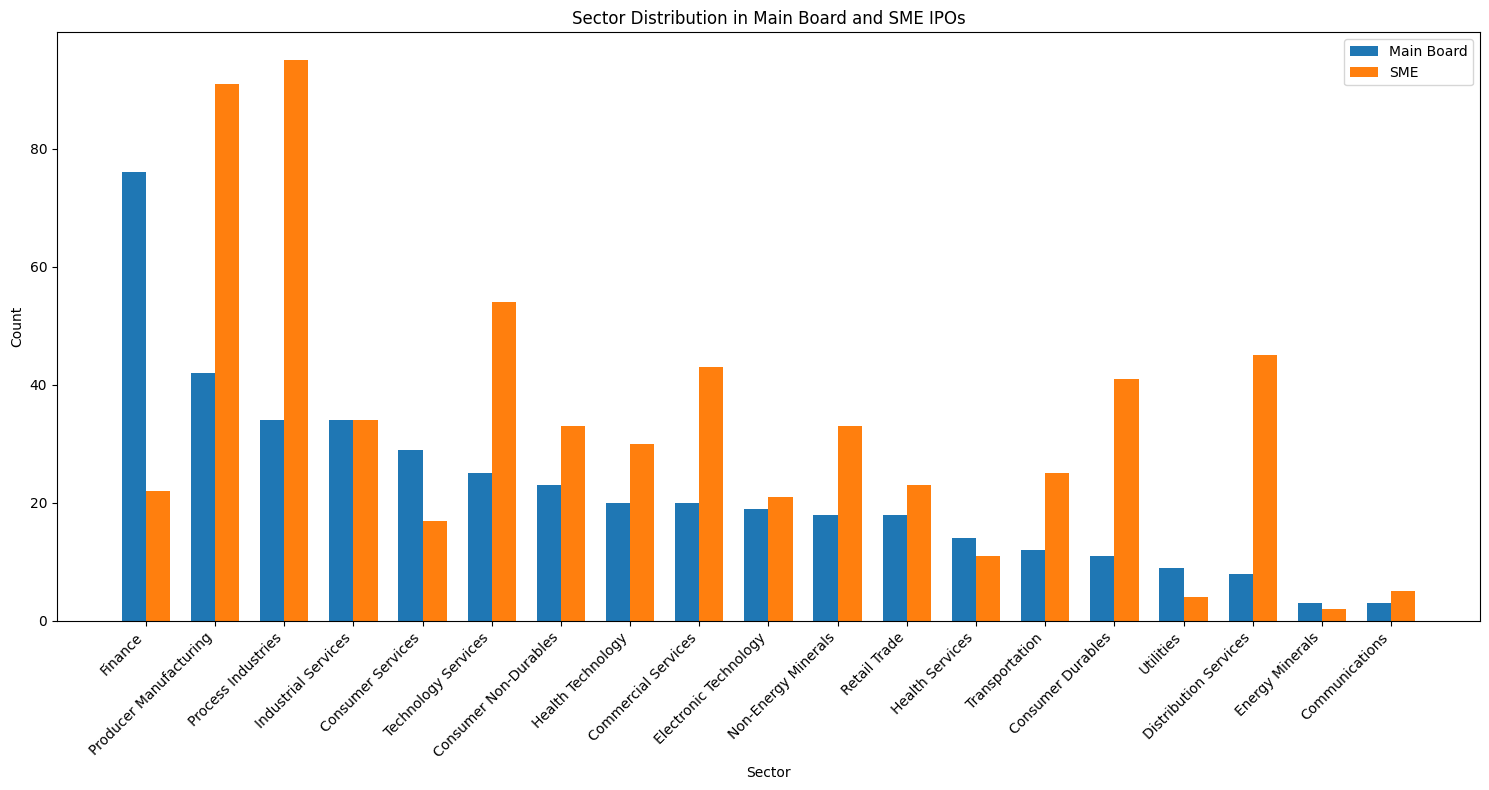}
\caption{Sector wise Distribution}
\label{fig:sector-dist}
\end{figure}

\begin{figure}[!ht]
\centering
\includegraphics[width=\textwidth]{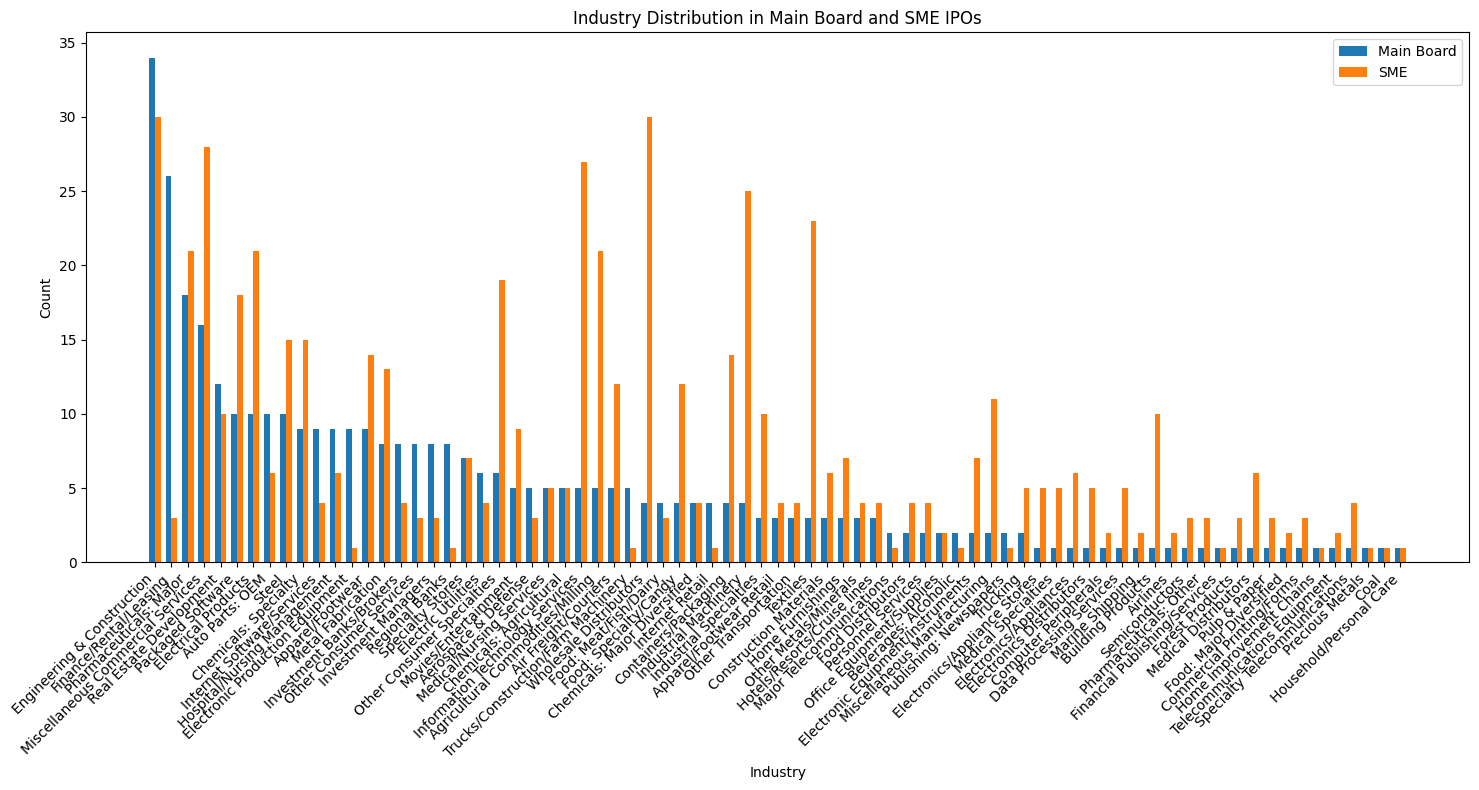}
\caption{Industry wise Distribution}
\label{fig:industry-dist}
\end{figure}

Subsequently, we collected historical values of Indian stock market indices i.e. Nifty 50 and Nifty VIX at daily, weekly, and monthly granularities from investing.com \footnote{
\url{https://in.investing.com/indices/s-p-cnx-nifty-historical-data?end_date=1714933800&interval_sec=weekly&st_date=1136053800&interval_sec=daily}}. We further extracted news articles related to the IPOs of the companies from Economic Times news portal. \footnote{\url{https://economictimes.indiatimes.com/archive.cms}} We could get 215 and 33 news articles which presented information regarding companies participating in Main Board and SME IPO respectively. In order to maintain data quality, we considered only those news articles which covered a single company. Furthermore, to get an understanding of the effect of various macroeconomic factors, we added them as features to our dataset. These features are: GDP per capita growth (annual),	GDP growth (annual),	GDP (current)	Unemployment rate,	stocks traded value,	Personal remittances, net trade in goods and services,	GNI per capita growth,	Inflation, consumer prices,	GNI (current),	Foreign direct investment. For a given IPO, we obtained the values of these features from the World Bank's website. \footnote{\url{https://data.worldbank.org/country/IN} (accessed on 25\textsuperscript{th} June, 2024)}. We obtained information regarding the sector and industry of the organization from stocksonfire.in \footnote{\href{https://stocksonfire.in/trading-ideas/nse-stocks-sector-wise-sorting-excel-sheet/}{https://stocksonfire.in/trading-ideas/nse-stocks-sector-wise-sorting-excel-sheet/} (accessed on 25\textsuperscript{th} June, 2024)}.  We present the sector wise and industry wise distributions in Figures \ref{fig:sector-dist} and \ref{fig:industry-dist} respectively. %To comprehend the market premium wherever possible, from chittorgarh.com we have obtained ratings of the IPOs from top brokers, analysts, and members. 
We engineered some features. They are the success rate of the IPOs launched in the previous quarter, and within the last 90 days from the launch of a given IPO. Information regarding the financials of the company, subscription rate till the penultimate day for subscription, ownership, ratings were obtained from chittorgarh.com \footnote{\url{https://www.chittorgarh.com/ipo/ipo_dashboard.asp} (accessed on 25\textsuperscript{th} June, 2024)}, and DHRP, RHP reports present in the SEBI \footnote{\url{https://www.sebi.gov.in/} (accessed on 25\textsuperscript{th} June, 2024)}, National Stock Exchange (NSE) \footnote{\url{https://www.nseindia.com/} (accessed on 25\textsuperscript{th} June, 2024)}, and Bombay Stock Exchange (BSE) \footnote{\url{https://www.bseindia.com/} (accessed on 25\textsuperscript{th} June, 2024)} websites. Subscription rate of a day is declared after the market closes on that day. We considered subscription rate till the penultimate day for subscription, as we want the investors to make a decision to opt for the IPO on the final day of subscription. 
In addition to this, we extracted texts, tables, and images from the prospectus (RHP, DRHP) of the companies. For these images, we performed Optical Character Recognition (OCR) using Tesseract \footnote{\url{https://github.com/tesseract-ocr/tesseract} (accessed on 25\textsuperscript{th} June, 2024)} to retrieve texts. For a given company, we stored the extracted content in a JavaScript Object Notation (JSON) file, where the keys corresponded to the pages in the prospectus. We complied a list of twenty-five questions which investors look for in the prospectus. These are presented in Table \ref{tab:questions} of \S\ref{appendix-questions}. These questions were formulated after interviewing several seasoned IPO investors and referring to eight reputed financial web-sites. \footnote{\href{https://www.motilaloswal.com/article-details/what-is-a-draft-red-herring-prospectus-and-why-is-it-important-for-investors/5259}{https://www.motilaloswal.com/article-details/what-is-a-draft-red-herring-prospectus-and-why-is-it-important-for-investors/5259}\\ \href{https://www.fisdom.com/what-to-look-for-in-an-rhp-before-investing-in-ipo/}{https://www.fisdom.com/what-to-look-for-in-an-rhp-before-investing-in-ipo/}\\ \href{https://www.indiainfoline.com/knowledge-center/ipo/what-is-a-draft-red-herring-prospectus}{https://www.indiainfoline.com/knowledge-center/ipo/what-is-a-draft-red-herring-prospectus}\\ \href{https://www.nism.ac.in/2024/01/understanding-drhp-rhp-and-prospectus/}{https://www.nism.ac.in/2024/01/understanding-drhp-rhp-and-prospectus/}\\ \href{https://groww.in/blog/things-you-must-know-about-rhp}{https://groww.in/blog/things-you-must-know-about-rhp} \href{https://www.chittorgarh.com/book-chapter/ipo-prospectus/18/}{https://www.chittorgarh.com/book-chapter/ipo-prospectus/18/}} \footnote{ \href{https://www.5paisa.com/stock-market-guide/ipo/things-to-know-in-rhp}{https://www.5paisa.com/stock-market-guide/ipo/things-to-know-in-rhp} \href{https://www.kotaksecurities.com/articles/6-things-to-look-for-in-a-draft-red-herring-prospectus/}{https://www.kotaksecurities.com/articles/6-things-to-look-for-in-a-draft-red-herring-prospectus/}  (accessed on 7\textsuperscript{th} September, 2024)} For each JSON file, we transformed the content of each page into embeddings using Nomic \cite{nussbaum2024nomic}. Nomic has a 8192 context length text encoder. Similarly, using Nomic we transformed each curated question to embeddings. From a given prospectus of a company, we retrieved pages relevant to the curated questions using cosine similarity and BM25 \cite{bm25s}. Cosine similarity was used for semantic matching, whereas BM25 was used for syntactic matching. Subsequently, we passed the retrieved pages and the corresponding question to a LLM, Llama3.2 -3b \cite{llama3modelcard} to generate the final answer. This approach is widely known as
Retrieval-Augmented Generation (RAG). The RAG based system treats each page of the PDF as a separate chunk.

Finally, for comparison, we obtained the GMP of the companies from investorgrain.com \footnote{\url{https://www.investorgain.com/report/live-ipo-gmp/331/} (accessed on 25\textsuperscript{th} June, 2024)}. We could not use GMP as a feature because we could get only the GMP values of the IPOs which were launched in the year 2019 or later. The final list of features and their descriptions are presented in Table \ref{tab:ipo-variables} of  \S\ref{appendix-variables}.

\begin{table}[t]%% placement specifier
\centering%% For centre alignment of tabular.
\caption{Training Test Split}\label{tab:data-dist}
\begin{tabular}{lcr} \hline
\textbf{Type} & \multicolumn{1}{l}{\textbf{Train}} & \multicolumn{1}{l}{\textbf{Test}} \\ \hline
Main Board     & 361                                & 57                                \\
SME           & 498                                & 183 \\  \hline                            
\end{tabular}
\end{table}

\section{ Experiments and results}
\label{sec:experiments}

\begin{figure}[t]%% placement specifier
%% Use \includegraphics command to insert graphic files. Place graphics files in 
%% working directory.
\centering%% For centre alignment of image.
\includegraphics[width=\columnwidth]{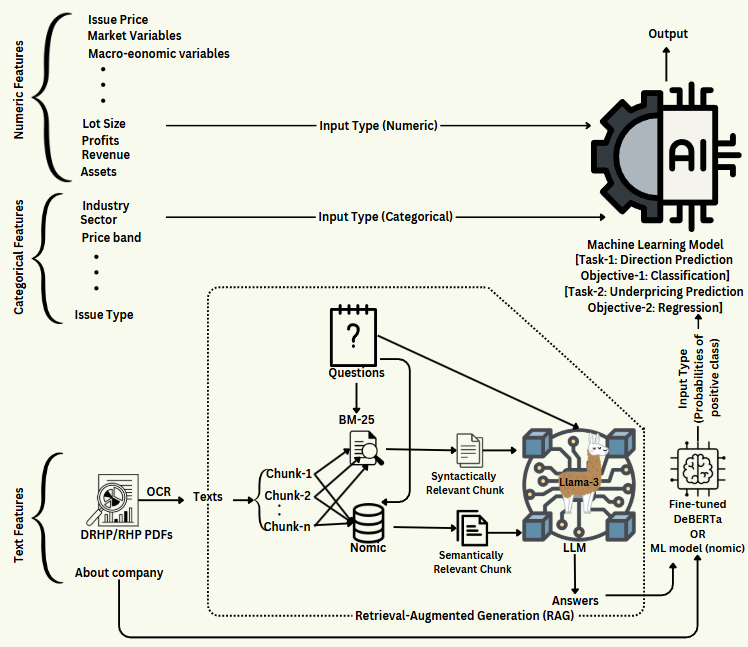}
%% Use \caption command for figure caption and label.
\caption{Methodology}
\label{fig:ipo-methodology}
\end{figure}

\begin{figure}[!ht]%% placement specifier
%% Use \includegraphics command to insert graphic files. Place graphics files in 
%% working directory.
\centering%% For centre alignment of image.
\includegraphics[]{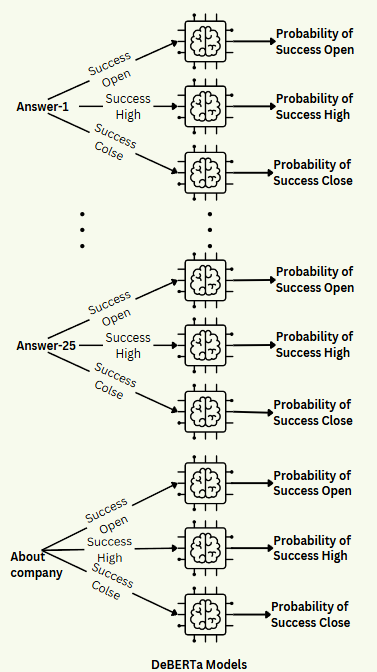}
%% Use \caption command for figure caption and label.
\caption{DeBERTa models}
\label{fig:ipo-deberta-models}
\end{figure}

We present the overall experimental framework in Figure \ref{fig:ipo-methodology}.

\subsection{Predicting direction of opening, high, and close prices of the listing day}
Our objective is to train separate models for predicting the direction of Opening, High and Closing prices. We started with using only the numeric and categorical (N-C) features for prediction. These numeric features are: Issue Price, Lot Size, Market Variables (Nifty50, VIX), Macroeconomic variables (GDP, Stocks traded, Unemployment rate, etc.), Subscription rate per category (QIB, NII, Retail, etc.) up to the penultimate day for subscription, Success rate of the IPOs in the previous quarter and the last 90 days, recommendations by brokers and members, face value of a share, Shares and Amount allocated to per category (Retail, HNI, etc.), Assets, Revenue, Profit After Tax, Net Worth, Reserves and Surplus, Total Borrowing, and Total Income. A comprehensive list of the features in presented in Table \ref{tab:ipo-variables}. 
We used the AutoML open-source library developed by the H2O team\footnote{\url{https://docs.h2o.ai/h2o/latest-stable/h2o-docs/automl.html} (accessed on 8\textsuperscript{th} September, 2024)} to train five kinds of models: Generalized Linear Model (GLM), Distributed Random Forest (DRF) \cite{randomforest}, neural networks based Deep Learning (DL) models, XG-Boost (XGB) \cite{xgboost}, and Gradient Boosting Machine (GBM) \cite{gbm}. Subsequently, we ensembled (Ens) these models to get the final predictions.

Later on we used text content (T) related to the company, (i.e. columns `full\_text\_content' and answer\_of\_question\_n)   in the modelling process. Moreover, we concatenated news (Nw) content related to the company (i.e. column: news\_content) with the text content (T). We could not use news content as a separate feature because it was present in less than 50\% and 10\% instances for main board and SME respectively. To include the texts as features, we firstly extracted their embeddings separately using Nomic \cite{nussbaum2024nomic}. We used these embeddings only as input features to train five kinds of models (GLM, DRF, DL, XGB, and GBM) leveraging H2O AutoML library for classification. For each text feature, separate ML models were trained for predicting the direction of opening, high and closing prices. We selected the best model in each case and appended the probabilities of the positive class as features to the list of existing numeric and categorical features. Direction equals to 1 (i.e.  opening price on the listing day of the IPO
greater than the issue price of the IPO) is referred to as a positive class.

Furthermore, as depicted in Figure \ref{fig:ipo-deberta-models} we fine-tuned 26 DeBERTa \cite{he2020deberta} models for classification corresponding to each of the 26 text features (column: full\_text\_content, answer\_of\_question\_1 to 25). This process was repeated three times to obtain probability of the positive class i.e. direction of price movement with respect to opening, high, and closing prices. We use these probabilities as inputs to the final machine learning based model. We replaced the previously mentioned Nomic based probabilities with the probabilities of the positive class obtained by fine-tuning separate DeBERTa-base \cite{he2020deberta} models. Each of the DeBERRTa models was trained for three epochs with learning rate of 2e-5 and batch size of 8. At a time, based on the objective, we use one out of these three models. This means for predicting success in terms of direction or under-pricing with respect to the opening price, we use the corresponding DeBERTa model which was fine-tuned for classifying direction of opening price.

The models for MB and SME IPOs were trained independently. For bench marking, we prompted a Gemini-1.5-flash \cite{reid2024gemini} model with all the necessary details for predicting the objectives. We repeated this with Llama-3.2 3b model. The details of the prompts are mentioned in section \ref{appendix-prompts}. We report Area Under the ROC curve (AUC), and F1 score for class 0 (i.e.,  F1(0)) and class 1 (i.e., F1(1)) in Table \ref{tab:classsification-results}.

Analysing the results, we observe that for predicting the direction of opening prices, Deep Learning (DL) and Gradient Boosting Machines (GBM) trained with numeric and categorical feature only performs the best for MB and SME respectively. However, for predicting direction of high prices, texts features do play a role. In this case, ML models trained with probability of positive class obtained by fine-tuning DeBERTa models as features dominate in terms of F1(1) for MB. But, in case of SME, XGB model trained with numerical, categorical features, and probabilities of positive class (obtained from the best performing ML models trained using Nomic embeddings) outperforms all others in terms of F1(1). Finally, for closing price, XGB models trained with numeric and categorical feature only performs the best in terms of F1(1) for both MB and SME.

% \textcolor{red}{CLARIFY WITH ARNAB} We removed all other columns which had more than 90\% instances as null. We create two other features: total subscription, and average subscription rate on the penultimate day. We imputed the missing values using Iterative Imputer.\footnote{\href{https://scikit-learn.org/stable/modules/impute.html}{https://scikit-learn.org/stable/modules/impute.html} (accessed on 8\textsuperscript{th} September, 2024)}

% We trained five kinds of machine learning based models for classification. These models are: Logistic Regression, Support Vector Machine \cite{svm}, Random Forest Classifier \cite{randomforest}, XG-Boost \cite{xgboost}, and LightGBM \cite{ke2017lightgbm}. These models were trained separately for the MB and SME IPOs as their characteristics are different. Subsequently, we converted the text column, `full\_text\_content'  into embeddings using Nomic \cite{nussbaum2024nomic} and used these embeddings as input features. We report Accuracy (Acc), and Weighted F1 (F1) score for evaluation in Table \ref{tab:classsification-results}.

%\textcolor{red}{MENTION, OTHER USED GLM and RANDOM FORESTS MODELS, WE FOUND OTHER MODELS PERFORMANCE BETTER/WORSE?}

% \usepackage{longtable}
% Note: It may be necessary to compile the document several times to get a multi-page table to line up properly

\begin{longtable}[c]{|lll|rrr|rrr|}
\caption{Results of predicting direction of opening, high, and close prices. O=Open, H=High, Cl=Close, N Numeric, C= Categorical, T = Raw Texts, Nw = News, Tn = Text Embeddings Probability (Nomic), Td = Text Embeddings Probability (DeBERTa), Llama = Llama 3.2 3b, Ens = Ensemble.  The best model (highest AUC, F1) of each type is highlighted in bold.}
\label{tab:classsification-results}\\
\hline
\multicolumn{3}{|l|}{}                                                                     & \multicolumn{3}{c|}{\textbf{Main Board}}                                                                          & \multicolumn{3}{c|}{\textbf{SME}}                                                                                 \\ \hline
\endfirsthead
\endhead
\multicolumn{1}{|l|}{\textbf{Model}} & \multicolumn{1}{l|}{\textbf{Type}} & \textbf{Input} & \multicolumn{1}{l|}{\textbf{AUC}}   & \multicolumn{1}{l|}{\textbf{F1 (0)}} & \multicolumn{1}{l|}{\textbf{F1 (1)}} & \multicolumn{1}{l|}{\textbf{AUC}}   & \multicolumn{1}{l|}{\textbf{F1 (0)}} & \multicolumn{1}{l|}{\textbf{F1 (1)}} \\ \hline

\multicolumn{1}{|l|}{GLM}            & \multicolumn{1}{l|}{O}             & N-C         & \multicolumn{1}{r|}{0.824}          & \multicolumn{1}{r|}{0.517}           & 0.915                                & \multicolumn{1}{r|}{\textbf{0.698}} & \multicolumn{1}{r|}{0.076}           & 0.887                                \\ \hline
\multicolumn{1}{|l|}{DRF}            & \multicolumn{1}{l|}{O}             & N-C         & \multicolumn{1}{r|}{0.597}          & \multicolumn{1}{r|}{0.592}           & 0.893                                & \multicolumn{1}{r|}{0.669}          & \multicolumn{1}{r|}{0.688}           & 0.890                                \\ \hline
\multicolumn{1}{|l|}{DL}             & \multicolumn{1}{l|}{O}             & N-C         & \multicolumn{1}{r|}{\textbf{0.903}} & \multicolumn{1}{r|}{\textbf{0.781}}  & \textbf{0.947}                       & \multicolumn{1}{r|}{0.679}          & \multicolumn{1}{r|}{0.003}           & 0.887                                \\ \hline
\multicolumn{1}{|l|}{XGB}       & \multicolumn{1}{l|}{O}             & N-C         & \multicolumn{1}{r|}{0.773}          & \multicolumn{1}{r|}{0.233}           & 0.893                                & \multicolumn{1}{r|}{0.670}          & \multicolumn{1}{r|}{0.077}  & 0.887                                \\ \hline
\multicolumn{1}{|l|}{GBM}            & \multicolumn{1}{l|}{O}             & N-C         & \multicolumn{1}{r|}{0.712}          & \multicolumn{1}{r|}{0.597}           & 0.893                                & \multicolumn{1}{r|}{0.606}          & \multicolumn{1}{r|}{\textbf{0.736}}           & \textbf{0.893}                       \\
\hline

\end{longtable}

\begin{longtable}{|lll|rr|rr|}
\caption{Results of predicting under pricing with respect opening, high, and close prices.O=Open, H=High, Cl=Close, N Numeric, C= Categorical, T = Raw Texts, Nw = News, Tn = Text Embeddings Probability (Nomic), Td = Text Embeddings Probability (DeBERTa), Ens = Ensemble. Best model (lowest MAE, MSE) of each type is highlighted in bold.}
\label{tab:regression-results}\\
\hline
\multicolumn{3}{|l|}{}                                                                     & \multicolumn{2}{c|}{\textbf{Main Board}}                                & \multicolumn{2}{c|}{\textbf{SME}}                                       \\ \hline
\endfirsthead
\endhead
\multicolumn{1}{|l|}{\textbf{Model}} & \multicolumn{1}{l|}{\textbf{Type}} & \textbf{Input} & \multicolumn{1}{l|}{\textbf{MAE}}   & \multicolumn{1}{l|}{\textbf{MSE}} & \multicolumn{1}{l|}{\textbf{MAE}}   & \multicolumn{1}{l|}{\textbf{MSE}} \\ \hline
\multicolumn{1}{|l|}{GLM}            & \multicolumn{1}{l|}{O}             & N-C            & \multicolumn{1}{r|}{0.208}          & 0.079                             & \multicolumn{1}{r|}{0.264}          & 0.136                             \\ \hline
\multicolumn{1}{|l|}{DRF}            & \multicolumn{1}{l|}{O}             & N-C            & \multicolumn{1}{r|}{0.247}          & 0.131                             & \multicolumn{1}{r|}{0.265}          & 0.169                             \\ \hline
\multicolumn{1}{|l|}{DL}             & \multicolumn{1}{l|}{O}             & N-C            & \multicolumn{1}{r|}{0.184}          & 0.066                             & \multicolumn{1}{r|}{0.248} & 0.152                             \\ \hline
\multicolumn{1}{|l|}{XGB}            & \multicolumn{1}{l|}{O}             & N-C            & \multicolumn{1}{r|}{0.231}          & 0.084                             & \multicolumn{1}{r|}{0.282}          & 0.157                             \\ \hline
\multicolumn{1}{|l|}{GBM}            & \multicolumn{1}{l|}{O}             & N-C            & \multicolumn{1}{r|}{0.223}          & 0.100                             & \multicolumn{1}{r|}{0.291}          & 0.214                             \\ \hline
\multicolumn{1}{|l|}{Ens}            & \multicolumn{1}{l|}{O}             & N-C            & \multicolumn{1}{r|}{0.171} & 0.057                    & \multicolumn{1}{r|}{0.248} & \textbf{0.127}                    \\ \hline

\multicolumn{1}{|l|}{GLM}            & \multicolumn{1}{l|}{O}             & N-C-Tn          & \multicolumn{1}{r|}{0.202}          & 0.083                             & \multicolumn{1}{r|}{0.264}          & 0.133                    \\
\hline
\end{longtable}

% \textcolor{red}{Add more description with respect to experiments related to text features}

% \textcolor{red}{Consult with Arnab, how text features and which text columns were used, whether PCA was done or only embeddings were used?}

\subsection{Predicting under-pricing with respect to opening, high, and close prices of the listing day}
Our objective is to train separate models for predicting underpricing with respect to Opening, High and Closing prices. Wherever we have these prices available in both BSE and NSE, we preferred to use the NSE prices. Similar to the previous section, we initiated the experiments with only the numeric features for prediction and added text features later. We trained the same five kind of machine learning models described previously for regression. Everything else other than the objective was kept the same. Our evaluation metrics were: Mean Squared Error (MSE) and Mean Absolute Error (MAE). We present the results in Table \ref{tab:regression-results}. We also prompted Llama 3.2 3b model under zero shot setting for predicting underpricing of MB IPOs. It could not predict the underpricing with respect to opening, high and closing prices for 36.84\%, 17.54\%, and  22.81\% cases respectively. Thus, we did not repeat this experiment with Llama 3.2 3b model to predict underpricing for the SME IPOs.

We observed that for predicting underpricing of SME IPOs with respect to opening price, the Ensemble model trained using numerical inputs, categorical inputs, and probabilities of positive class as features performed the best in terms of MAE. These probabilities of positive class were obtained from best performing ML models trained using Nomic embeddings of text columns for predicting the direction of opening prices. However, in case of predicting underpricing of MB IPOs with respect to Opening price, Deep Learning (DL) model trained using numerical, categorical inputs, and probabilities obtained by fine-tuning DeBERTa models performed the best in terms of MAE.

Similarly, for predicting underpricing with respect to high prices, Ensemble and DL models trained with probabilities of positive classes along with numeric and categorical features performed the best in terms of MAE for MB and SME respectively. As mentioned before, the probabilities of positive class were obtained from the best performing ML models trained using Nomic embeddings of text columns for predicting the direction of high prices. We also observed that in case of MB, News content (Nw) played a role.

 Finally, for predicting underpricing with respect to closing prices, DL models trained with probabilities of positive classes along with numeric and categorical features performed the best in terms of MAE for both MB and SME. As discussed previously, the probabilities of positive class were obtained from the best preforming ML models trained using Nomic embeddings of text columns for predicting the direction of closing prices.

\subsection{Experiments related to Grey Market Premium}
To understand the relation between GMP and success of the IPOs, we collected GMP of the 287 and 385 companies which went for IPOs in Main Board and SME respectively. In Table \ref{tab:gmp}, we present how the listing price and issue price varied when GMP was negative, zero, and positive. We considered all the companies when went for Main Board or SME IPO from 1\textsuperscript{st} January 2019 to 12\textsuperscript{th} July 2024. We eliminated those cases where GMP were not available. For the year, 2023 we present the results separately as it corresponds to the test set on which we are doing all our evaluation. It is interesting to note that at an overall level, GMP values aligned with the difference between Listing Prices and Issue Prices in 80.29\% cases for the Main Board IPOs and 21.29\% cases for the SME IPOs.

Using GMP, we predicted the under-pricing with respect to opening price on the listing day i.e. ((GMP + issue price) - issue price)/(issue price) = (GMP)/(issue price). We compared it with the actual under-pricing, i.e. (opening price - issue Price)/(issue price). For the entire main board dataset, we obtained 
MAE, and MSE as 0.109, and 0.031 respectively. For the year 2023 only, the values of MAE, and MSE for main board IPOs are 0.091, and 0.019 respectively. Similarly, for the entire SME dataset, we obtained MAE, and MSE as 0.751, and 1.509 respectively. For the year 2023 only, the values of MAE, and MSE for SME IPOs are 0.531, and 0.771 respectively. Based on availability of data in investorgain.com, we present this analysis with respect to opening price only. 

We observe that GMP does a good job for predicting success of main board IPOs. However, for SME IPOs, GMP is not a good indicator.

\begin{table}[h]
\centering
\caption{GMP analysis. IP = Issue Price, LP = Listing Price.}\label{tab:gmp}
\begin{tabular}{|ll|rrr|rrr|}
\hline
\multicolumn{2}{|l|}{\multirow{2}{*}{\textbf{}}}                                       & \multicolumn{3}{c|}{\textbf{GMP (overall)}}                                                         & \multicolumn{3}{c|}{\textbf{GMP (2023)}}                                                            \\ \cline{3-8} 
\multicolumn{2}{|l|}{}                                                                 & \multicolumn{1}{l|}{\textless{}0} & \multicolumn{1}{l|}{=0} & \multicolumn{1}{l|}{\textgreater{}0} & \multicolumn{1}{l|}{\textless{}0} & \multicolumn{1}{l|}{=0} & \multicolumn{1}{l|}{\textgreater{}0} \\ \hline
\multicolumn{1}{|l|}{\multirow{3}{*}{\textbf{Main Board}}} & LP\textless{}IP    & \multicolumn{1}{r|}{18}           & \multicolumn{1}{r|}{5}   & 15                                   & \multicolumn{1}{r|}{0}            & \multicolumn{1}{r|}{2}   & 3                                    \\ \cline{2-8} 
\multicolumn{1}{|l|}{}                                            & LP=IP              & \multicolumn{1}{r|}{1}            & \multicolumn{1}{r|}{0}   & 3                                    & \multicolumn{1}{r|}{0}            & \multicolumn{1}{r|}{0}   & 1                                    \\ \cline{2-8} 
\multicolumn{1}{|l|}{}                                            & LP\textgreater{}IP & \multicolumn{1}{r|}{9}            & \multicolumn{1}{r|}{8}   & 149                                  & \multicolumn{1}{r|}{0}            & \multicolumn{1}{r|}{3}   & 50                                   \\ \hline
\multicolumn{1}{|l|}{\multirow{3}{*}{\textbf{SME}}}        & LP\textless{}IP    & \multicolumn{1}{r|}{16}           & \multicolumn{1}{r|}{42}  & 232                                  & \multicolumn{1}{r|}{14}           & \multicolumn{1}{r|}{9}   & 98                                   \\ \cline{2-8} 
\multicolumn{1}{|l|}{}                                            & LP=IP              & \multicolumn{1}{r|}{3}            & \multicolumn{1}{r|}{6}   & 20                                   & \multicolumn{1}{r|}{2}            & \multicolumn{1}{r|}{5}   & 13                                   \\ \cline{2-8} 
\multicolumn{1}{|l|}{}                                            & LP\textgreater{}IP & \multicolumn{1}{r|}{1}            & \multicolumn{1}{r|}{5}   & 60                                   & \multicolumn{1}{r|}{1}            & \multicolumn{1}{r|}{3}   & 34                                   \\ \hline
\end{tabular}
\end{table}

\section{Conclusion}
\label{sec:conclusion}
% Discussion and conclusion
% Future works: Expert review and rating generate automatically, social media hype (market premium) capture
%\textcolor{red}{Industry wise performance}

%\textcolor{magenta}{Paraphrase and re-write our contributions}

% {\textcolor{violet}{a short speech that is given to encourage someone to work harder, to feel more confident and enthusiastic, etc.}}

In this paper, we thoroughly studied the Indian IPO landscape separately for main board and SME listed companies. We curated two new datasets. We mined relevant information from DRHP and RHP reports. We examined how different macroeconomic factors, the current performance of the stock market, and a company's financial health influence the success of its initial public offering (IPO). We also observed that, GMP can be used as a proxy for estimating success of Main Board IPOs. However, for SME IPOs, GMP is not a good indicator.

We experimented with multi-classifier decision system that fuses information from different modalities. We used texts, images, numerical data, and categorical features as inputs to predict the direction and underpricing of stock prices at the opening, high, and closing points on the IPO listing day. For predicting the direction of opening and closing prices, ML models like Deep Learning (DL), XG-Boost (XGB) and Gradient Boosting Machines (GBM) demonstrate superior performance when trained with numeric and categorical features. However, when it comes to predicting direction of high prices with respect to the issue price of an IPO, incorporating text features enhances performance of the predictors, particularly with prediction probabilities from DeBERTa models and ML models trained using Nomic embeddings. Interestingly, our approaches outperformed Gemini 1.5 flash and Llama 3.2 3b (a popular large language models) under zero shot setting. For underpricing predictions with respect to opening price, the Ensemble model excels for SME IPOs when leveraging a combination of numerical, categorical inputs, and probabilities derived from ML models. Conversely, the DL model is more effective for MB IPOs under similar conditions. This trend continues for under-pricing predictions with respect to high and closing price, where both Ensemble and DL models trained with a blend of feature types consistently yield the best results. In summary, our analysis reveals that the effectiveness of various machine learning models in predicting IPO price movements.

Overall, our findings underscore the importance of feature engineering in enhancing prediction accuracy in IPO pricing. This highlights the potential of advanced machine learning techniques to leverage both structured and unstructured data effectively.

This study has some limitations which can be the grounds for future works. Firstly, we have not considered the market premium which gets created due to discussions in social media. We could not consider the reviews written by expert analysts about the IPOs because this was only available from the year 2016 onwards. In future, we would like to extensively work in mining these reviews. Secondly, changing regulations can have effect on the success of an IPO. Although an IPO is regulated by SEBI, there can be instances of manipulating the Earnings Per Share (EPS)  of a company before IPO. Impact of various local and international events like COVID-19, Russia Ukraine war, and General Elections of India in 2009, 2014, 2019 factors were not considered. We could gather information related to IPOs for 1099 instances in total. Expanding this dataset and capturing more features related to stock market dynamics are directions for further research. Furthermore, the models we proposed depend on historical performance data, which may not consistently reflect future results. Other future research directions include investigating how investor behaviour and psychological factors influence IPO pricing and performance, which could yield valuable insights into market dynamics. Employing qualitative methodologies such as interviews and case studies would enhance the understanding of the IPO process from the perspectives of both issuers and investors. Additionally, a comparative analysis with other emerging markets could provide broader context and insights. Further research exploring the impact of emerging technologies, and specific industry characteristics, will offer valuable insights for enhancing the efficiency and sustainability of the Indian IPO market. Currently, we have extracted textual data from the images contained within the prospectus. A more effective approach would involve the utilization of multi-modal embeddings. Similarly, various chunking strategies can be used to evaluate and improve the RAG system. Finally, exploring the impact of non-financial factors, including corporate governance and social responsibility, on IPO pricing could deepen the understanding of the IPO market dynamics.

\section*{CRediT authorship contribution statement}
\textbf{Sohom Ghosh:} Conceptualization, Methodology, Software, Validation, Formal analysis, Investigation, Resources Data Curation, Writing - Original Draft, Visualization, Project administration.
\textbf{Arnab Maji:} Software, Validation, Formal analysis, Resources Data Curation.
\textbf{N Harsha Vardhan:} Software, Validation, Formal analysis, Resources Data Curation.
\textbf{Sudip Kumar Naskar:} Writing: Review \& Editing, Supervision.
%\url{https://www.elsevier.com/en-in/researcher/author/policies-and-guidelines/credit-author-statement}

\section*{Declaration of competing interest}
The authors declare that they have no known competing financial interests or personal relationships that could have appeared to influence the work reported in this paper.

\section*{Data availability}
The datasets can be downloaded from 
HuggingFace \url{https://huggingface.co/datasets/sohomghosh/Indian_IPO_datasets}.

\section*{Reproducibility}
For ensuring our work is reproducible, we have provided all the necessary details, including the hyper-parameters corresponding to the best performing models in GitHub \url{https://github.com/sohomghosh/Indian_IPO}.

\section*{Acknowledgments}
We would like to thank Souvik Meta and Aswartha Narayana for their inputs regarding pricing of IPOs and features effecting its success.

\bibliographystyle{elsarticle-harv} 
\bibliography{bibliography.bib}




% %% For authoryear reference style

% \bibitem[Lamport(1994)]{lamport94}
%   \textit{\LaTeX: a document preparation system},
%   Addison Wesley, Massachusetts,

% \end{thebibliography}

% \newpage
\appendix
\section{Questions}
\label{appendix-questions}

A list of curated questions is presented in Table \ref{tab:questions}.

\begin{table}[ht]
\caption{List of Questions}
\label{tab:questions}
\resizebox{\textwidth}{!}{%
\begin{tabular}{ll}
\hline
\textbf{no} & \textbf{question }      \\    \hline   %\\
1  & \begin{tabular}[c]{@{}l@{}}What are the background, qualifications, and experience of the promoters\\ and key management team?\end{tabular}                                                                 \\
2  & \begin{tabular}[c]{@{}l@{}}Are there any past or current criminal cases, police cases, or legal  \\ proceedings legal cases against the promoters and \\ key management team or the   company?\end{tabular} \\
3  & \begin{tabular}[c]{@{}l@{}}Are the promoters and key management team capable of managing\\ the company   and meet its objectives?\end{tabular}                                                              \\
4  & What are the company's business model, products/services?                                                                                                                                                   \\
5  & \begin{tabular}[c]{@{}l@{}}What are the company's strengths, competitive advantages, and \\ growth   potential?\end{tabular}                                                                                \\
6  & How is the company's position in the industry?                                                                                                                                                              \\
7  & Is the company able to adapt to market changes?                                                                                                                                                             \\
8  & What are the company's growth prospects?                                                                                                                                                                    \\
9  & \begin{tabular}[c]{@{}l@{}}What is the financial performance of the company in terms of \\ revenue,   profits, assets, and liabilities?\end{tabular}                                                        \\
10 & How does the company plan to use the funds raised through IPO?                                                                                                                                              \\
11 & Does the company plan to use IPO proceeds to repay debt?                                                                                                                                                    \\
12 & \begin{tabular}[c]{@{}l@{}}Does the proposed use of funds raised from IPO aligns \\ with the company's   growth strategy?\end{tabular}                                                                      \\
13 & \begin{tabular}[c]{@{}l@{}}What is the potential impact of the IPO on the \\ company's future prospects?\end{tabular}                                                                                       \\
14 & What are the risks associated with investing in the company?                                                                                                                                                \\
15 & Is the company's able to mitigate the risks and their potential impact?                                                                                                                                     \\
16 & \begin{tabular}[c]{@{}l@{}}What is the potential impact of market fluctuations \\ on the company's performance?\end{tabular}                                                                                \\
17 & Is the IPO price is reasonable and offers potential for growth?                                                                                                                                             \\
18 & \begin{tabular}[c]{@{}l@{}}Does the IPO price reflect the company's intrinsic value \\ and growth prospects?\end{tabular}                                                                                   \\
19 & Is the IPO price is reasonable and offers potential for growth?                                                                                                                                             \\
20 & \begin{tabular}[c]{@{}l@{}}Is the company's valuation right based on financial metrics \\ (like P/E   ratio, Enterprise value-to-EBITDA ratio) \\ and industry comparisons?\end{tabular}                    \\
21 & How is the company's position in the market and among its competitors?                                                                                                                                      \\
22 & Has the company complied with all relevant regulatory requirements?                                                                                                                                         \\
23 & Who are the lead, and co-lead managers/under-writers?                                                                                                                                                       \\
24 & \begin{tabular}[c]{@{}l@{}}What is the company's corporate governance structure, including board \\ composition, executive compensation, and shareholder rights.?\end{tabular}                              \\
25 & \begin{tabular}[c]{@{}l@{}}What is the shareholding pattern, i.e.the ownership structure \\ and potential conflicts of interest?\end{tabular} \\                         \hline                                    
\end{tabular}%
}
\end{table}

\section{Variables}
\label{appendix-variables}
A list of variables which are used for predicting success of IPOs are presented in Table \ref{tab:ipo-variables}.

% \usepackage{lscape}
% \usepackage{longtable}
% Note: It may be necessary to compile the document several times to get a multi-page table to line up properly
\begin{landscape}
\begin{longtable}[c]{|l|l|l|l|}
\caption{Description of Variables. P = Presence (B= Both, M = Main Board, S = SME),	T = Type of variable (I = Independent Variable i.e. Features, D = Dependent Variable i.e. Target)}
\label{tab:ipo-variables}\\
\hline
\textbf{P} & \textbf{T} & \textbf{Column Name}                                                                                               & \textbf{Description}                                                                                                                              \\ \hline
\endfirsthead
\multicolumn{4}{c}%
{{\bfseries Table \thetable\ continued from previous page}} \\
\hline
\textbf{P} & \textbf{T} & \textbf{Column Name}                                                                                               & \textbf{Description}                                                                                                                              \\ \hline
\endhead
B          & I          & mapping\_key                                                                                                       & Unique key for identifying each IPO                                                                                                               \\ \hline
B          & I          & Company Name                                                                                                       & Name of the company going for IPO in short                                                                                                        \\ \hline
B          & I          & Issuer Company                                                                                                     & Full Name of the company going for IPO                                                                                                            \\ \hline
B          & I          & url                                                                                                                & URL corresponding to the company's IPO in chittorgarh.com                                                                                         \\ \hline
B          & I          & subscription\_link                                                                                                 & URL to access the subscription information                                                                                                        \\ \hline
M          & I          & Subscription Dates                                                                                                 & Dates on which IPO can be subscribed                                                                                                              \\ \hline
B          & I          & NSE\_symbol                                                                                                        & Ticker of the company in NSE                                                                                                                      \\ \hline
B          & I          & Total Issue Size                                                                                                   & Total monetary value of all shares being offered to the public                                                                                    \\ \hline
B          & I          & Offer for Sale                                                                                                     & Value of shares being offered for sale                                                                                                            \\ \hline
B          & I          & Issue Type                                                                                                         & Fixed Price Issue or Book Building Issue                                                                                                          \\ \hline
B          & I          & Listing Date                                                                                                       & Day on which IPO will get listed                                                                                                                  \\ \hline
B          & I          & Price Band                                                                                                         & The range of prices within which investors can bid for shares                                                                                     \\ \hline
B          & I          & Industry                                                                                                           & Industry of the company                                                                                                                           \\ \hline
B          & I          & Sector                                                                                                             & Sector of the company                                                                                                                             \\ \hline
B          & I          & IPO Date                                                                                                           & Duration for subscribing the IPO                                                                                                                  \\ \hline
B          & I          & Close Date                                                                                                         & Last day for IPO subscription                                                                                                                     \\ \hline
B          & I          & Close Year                                                                                                         & Year of the last day for IPO subscription                                                                                                         \\ \hline
B          & I          & Close Year Previous                                                                                                & One year before the year of the last day for IPO subscription                                                                                     \\ \hline
M          & I          & Exchange                                                                                                           & NSE or BSE                                                                                                                                        \\ \hline
M          & I          & Issue Size (Rs Cr.)                                                                                                & Total shares that a company proposes to offer                                                                                                     \\ \hline
B          & I          & Final\_Issue\_Price                                                                                                & \begin{tabular}[c]{@{}l@{}}Final Issue price of the company (on NSE for Main Board)\\ (on BSE/NSE for SME whichever available) \end{tabular}                                                                                                                 \\ \hline
S          & I          & BSE\_Final\_Issue\_Price                                                                                           & Issue price of the IPO on BSE                                                                                                                     \\ \hline
S          & I          & NSE\_Final\_Issue\_Price                                                                                           & Final Issue price of the SME IPO on NSE                                                                                                           \\ \hline
B          & I          & Fresh Issue                                                                                                        & Value of shares being freshly issued                                                                                                              \\ \hline
M          & I          & Lot Size                                                                                                           & Minimum number of shares that an investor must bid for                                                                                            \\ \hline
M          & I          & Open Date                                                                                                          & Date when IPO will be opened for subscription                                                                                                     \\ \hline
B          & I          & \textless{}n\textgreater{}\_assets                                                                                 & Asset of the company as on nth day                                                                                                                \\ \hline
B          & I          & \textless{}n\textgreater{}\_net worth                                                                              & Net worth of the company as on nth day                                                                                                            \\ \hline
B          & I          & \textless{}n\textgreater{}\_profit after tax                                                                       & Profit after tax of the company on the nth day                                                                                                    \\ \hline
B          & I          & \textless{}n\textgreater{}\_reserves and surplus                                                                   & Reserves and surplus of the company on the nth day                                                                                                \\ \hline
B          & I          & \textless{}n\textgreater{}\_revenue                                                                                & Revenue of the company on the nth day                                                                                                             \\ \hline
B          & I          & \textless{}n\textgreater{}\_total borrowing                                                                        & Total borrowing of the company on the nth day                                                                                                      \\ \hline
B          & I          & \textless{}n\textgreater{}\_total income                                                                           & Total income of the company on the nth day                                                                                                        \\ \hline
B          & I          & 1 day before Close Day                                                                                             & Date before 1 day before the last day for IPO subscription                                                                                        \\ \hline
B          & I          & 1 month before Close Day month number                                                                              & \begin{tabular}[c]{@{}l@{}}Month number of the month which is 1 month before\\  the last day of IPO subscription\end{tabular}                     \\ \hline
B          & I          & \begin{tabular}[c]{@{}l@{}}1 month before \\ Close Day year number\end{tabular}                                    & \begin{tabular}[c]{@{}l@{}}Year of the month which is 1 month before\\  the last day of IPO subscription\end{tabular}                            \\ \hline
B          & I          & \begin{tabular}[c]{@{}l@{}}1 week before \\ Close Day week number\end{tabular}                                     & \begin{tabular}[c]{@{}l@{}}Week number of the week which is 1 week before\\ the last day of IPO subscription\end{tabular}                         \\ \hline
B          & I          & \begin{tabular}[c]{@{}l@{}}1 week before \\ Close Day year number\end{tabular}                                     & \begin{tabular}[c]{@{}l@{}}Year of the week which is 1 week before the \\ last day of IPO subscription\end{tabular}                               \\ \hline
B          & I          & Basis of Allotment                                                                                                 & \begin{tabular}[c]{@{}l@{}}Date when the final allocation of shares in an \\ Initial Public Offering (IPO) is disclosed to investors\end{tabular} \\ \hline
B          & I          & Brokers\_Avoid                                                                                                     & \begin{tabular}[c]{@{}l@{}}Number of top Brokers \& Analysts who \\ recommended to avoid the IPO\end{tabular}                                     \\ \hline
B          & I          & Brokers\_Neutral                                                                                                   & \begin{tabular}[c]{@{}l@{}}Number of top Brokers \& Analysts with neutral \\ recommended for the IPO\end{tabular}                                 \\ \hline
B          & I          & Brokers\_Subscribe                                                                                                 & \begin{tabular}[c]{@{}l@{}}Number of top Brokers \& Analysts who recommended \\ to subscribe for the IPO\end{tabular}                             \\ \hline
B          & I          & Members\_Avoid                                                                                                     & Number of Members who recommended to avoid the IPO                                                                                                \\ \hline
B          & I          & Members\_Neutral                                                                                                   & Number of Members with neutral recommended for the IPO                                                                                            \\ \hline
B          & I          & Members\_Subscribe                                                                                                 & \begin{tabular}[c]{@{}l@{}}Number of Members who recommended to \\ subscribe for the IPO\end{tabular}                                             \\ \hline
B          & I          & Face Value per share                                                                                               & Face value of a share                                                                                                                             \\ \hline
B          & I          & Credit of Shares to Demat                                                                                          & Date on which shares would be credited to the demat account                                                                                       \\ \hline
B          & I          & \begin{tabular}[c]{@{}l@{}}Cut-off time for \\ UPI mandate confirmation\end{tabular}                               & Time by which an investor must approve the UPI mandate request                                                                                    \\ \hline
B          & I          & day\_\textless{}n\textgreater{}\_date                                                                              & Date of day n                                                                                                                                     \\ \hline
B          & I          & day\_\textless{}n\textgreater{}\_bNII (bids above 10L)                                                             & Subscriptions by big Non-Institutional Investors on day n                                                                                         \\ \hline
B          & I          & day\_\textless{}n\textgreater{}\_emp                                                                               & Subscriptions by employees on day n                                                                                                               \\ \hline
B          & I          & day\_\textless{}n\textgreater{}\_nii                                                                               & Subscriptions by Non-Institutional Investors (NII) on day n                                                                                       \\ \hline
B          & I          & day\_\textless{}n\textgreater{}\_nii*                                                                              & \begin{tabular}[c]{@{}l@{}}Subscriptions by other type of \\ Non-Institutional Investors (NII*) on day n\end{tabular}                             \\ \hline
B          & I          & day\_\textless{}n\textgreater{}\_other                                                                             & Subscriptions by others on day n                                                                                                                  \\ \hline
B          & I          & day\_\textless{}n\textgreater{}\_qib                                                                               & Subscriptions by Qualified Institutional Buyers (QIB) on day n                                                                                    \\ \hline
B          & I          & day\_\textless{}n\textgreater{}\_retail                                                                            & Subscriptions by Retail Investors on day n                                                                                                        \\ \hline
B          & I          & day\_\textless{}n\textgreater{}\_total                                                                             & Total subscriptions on day n                                                                                                                      \\ \hline
B          & I          & Retail (Max)\_Amount                                                                                               & Maximum application amount for Retailers                                                                                                          \\ \hline
B          & I          & Retail (Max)\_Lots                                                                                                 & Maximum number of lots that a Retailers must apply for                                                                                            \\ \hline
B          & I          & Retail (Max)\_Shares                                                                                               & Maximum number of shares that a Retailers must apply for                                                                                          \\ \hline
B          & I          & Retail (Min)\_Amount                                                                                               & Minimum application amount for Retailers                                                                                                          \\ \hline
B          & I          & Retail (Min)\_Lots                                                                                                 & Minimum number of lots that a Retailers must apply for                                                                                            \\ \hline
B          & I          & Retail (Min)\_Shares                                                                                               & Minimum number of shares that a Retailers must apply for                                                                                          \\ \hline
M          & I          & B-HNI (Min)\_Amount                                                                                                & \begin{tabular}[c]{@{}l@{}}Minimum application amount for Big High \\ Net-worth Individuals (B-HNIs)\end{tabular}                                 \\ \hline
M          & I          & B-HNI (Min)\_Lots                                                                                                  & Minimum number of lots that a B-HNI must apply for                                                                                                \\ \hline
M          & I          & B-HNI (Min)\_Shares                                                                                                & Minimum number of shares that a B-HNI must apply for                                                                                              \\ \hline
M          & I          & S-HNI (Max)\_Amount                                                                                                & \begin{tabular}[c]{@{}l@{}}Maximum application amount for Small High \\ Net-worth Individuals (S-HNIs)\end{tabular}                               \\ \hline
M          & I          & S-HNI (Max)\_Lots                                                                                                  & Maximum number of lots that a S-HNI must apply for                                                                                                \\ \hline
M          & I          & S-HNI (Max)\_Shares                                                                                                & Maximum number of shares that a S-HNI must apply for                                                                                              \\ \hline
M          & I          & S-HNI (Min)\_Amount                                                                                                & Minimum application amount for S-HNI                                                                                                              \\ \hline
M          & I          & S-HNI (Min)\_Lots                                                                                                  & Minimum number of lots that a S-HNI must apply for                                                                                                \\ \hline
M          & I          & S-HNI (Min)\_Shares                                                                                                & Minimum number of shares that a S-HNI must apply for                                                                                              \\ \hline
S          & I          & HNI (Min)\_Amount                                                                                                  & \begin{tabular}[c]{@{}l@{}}Minimum application amount for High \\ Net-worth Individuals (HNIs)\end{tabular}                                       \\ \hline
S          & I          & HNI (Min)\_Lots                                                                                                    & Minimum number of lots that a HNI must apply for                                                                                                  \\ \hline
S          & I          & HNI (Min)\_Shares                                                                                                  & Minimum number of shares that a HNI must apply for                                                                                                \\ \hline
B          & I          & Share Holding Post Issue                                                                                           & Distribution of ownership stakes in a company post IPO                                                                                            \\ \hline
B          & I          & Share Holding Pre Issue                                                                                            & Distribution of ownership stakes in a company pre IPO                                                                                             \\ \hline
B          & I          & Stocks traded, total value (\% of GDP)                                                                             & Value of stock traded as \% of GDP in the year prior to IPO                                                                                       \\ \hline
B          & I          & dhrp\_rhp\_links                                                                                                   & Links for DHRP, RHP, Anchor Investor files                                                                                                        \\ \hline
B          & I          & dhrp\_rhp\_links\_pdf                                                                                              & Links for DHRP, RHP, Anchor Investor pdf files                                                                                                    \\ \hline
B          & I          & most\_relevant\_link                                                                                               & \begin{tabular}[c]{@{}l@{}}Link to download prospectus. Preference is given \\ to RHP followed by DRHP and Anchor Investor files.\end{tabular}    \\ \hline
B          & I          & File\_Rename\_1st                                                                                                  & \begin{tabular}[c]{@{}l@{}}Name of downloaded PDF file. Preference is given \\ to RHP followed by DRHP and Anchor Investor files\end{tabular}     \\ \hline
B          & I          & Text\_extracted\_JSON                                                                                              & Page wise texts extracted from the PDF in JSON format                                                                                              \\ \hline
B          & I          & full\_text\_content                                                                                                & Text content related to the IPO obtained from chittorgarh.com                                                                                     \\ \hline
B          & I          & news\_content                                                                                                      & List of news relating to the company's IPO                                                                                                        \\ \hline
B          & I          & news\_headline                                                                                                     & List of news headlines relating to the company's IPO                                                                                              \\ \hline
B          & I          & news\_synopsis                                                                                                     & List of news synopsis relating to the company's IPO                                                                                               \\ \hline
B          & I          & news\_url                                                                                                          & \begin{tabular}[c]{@{}l@{}}List of URLs corresponding to the news \\ relating to the company's IPO\end{tabular}                                   \\ \hline
B          & I          & Chg\%\_nifty50\_daily                                                                                              & \begin{tabular}[c]{@{}l@{}}Change in nifty 50 index during the \\ day previous to the Close Date\end{tabular}                                     \\ \hline
B          & I          & Chg\%\_nifty50\_monthly                                                                                            & \begin{tabular}[c]{@{}l@{}}Change in nifty 50 index during the \\ month previous to the Close Date\end{tabular}                                   \\ \hline
B          & I          & Chg\%\_nifty50\_weekly                                                                                             & \begin{tabular}[c]{@{}l@{}}Change in nifty 50 index during the \\ week previous to the Close Date\end{tabular}                                    \\ \hline
B          & I          & Chg\%\_vix\_daily                                                                                                  & \begin{tabular}[c]{@{}l@{}}Change in vix index during the day \\ previous to the Close Date\end{tabular}                                          \\ \hline
B          & I          & Chg\%\_vix\_monthly                                                                                                & \begin{tabular}[c]{@{}l@{}}Change in vix index during the month \\ previous to the Close Date\end{tabular}                                        \\ \hline
B          & I          & Chg\%\_vix\_weekly                                                                                                 & \begin{tabular}[c]{@{}l@{}}Change in vix index during the week \\ previous to the Close Date\end{tabular}                                         \\ \hline
B          & I          & Open\_nifty50\_daily                                                                                               & \begin{tabular}[c]{@{}l@{}}Opening Price of nifty 50 index during \\ the day previous to the Close Date\end{tabular}                              \\ \hline
B          & I          & Open\_nifty50\_monthly                                                                                             & \begin{tabular}[c]{@{}l@{}}Opening Price of nifty 50 index during \\ the month previous to the Close Date\end{tabular}                            \\ \hline
B          & I          & Open\_nifty50\_weekly                                                                                              & \begin{tabular}[c]{@{}l@{}}Opening Price of nifty 50 index during \\ the week previous to the Close Date\end{tabular}                             \\ \hline
B          & I          & Open\_vix\_daily                                                                                                   & \begin{tabular}[c]{@{}l@{}}Opening Price of vix during the day \\ previous to the Close Date\end{tabular}                                         \\ \hline
B          & I          & Open\_vix\_monthly                                                                                                 & \begin{tabular}[c]{@{}l@{}}Opening Price of vix during the month \\ previous to the Close Date\end{tabular}                                       \\ \hline
B          & I          & Open\_vix\_weekly                                                                                                  & \begin{tabular}[c]{@{}l@{}}Opening Price of vix during the week \\ previous to the Close Date\end{tabular}                                        \\ \hline
B          & I          & High\_nifty50\_daily                                                                                               & \begin{tabular}[c]{@{}l@{}}Highest value of nifty 50 index during the \\ day previous to the Close Date\end{tabular}                              \\ \hline
B          & I          & High\_nifty50\_monthly                                                                                             & \begin{tabular}[c]{@{}l@{}}Highest value of nifty 50 index during the \\ month previous to the Close Date\end{tabular}                            \\ \hline
B          & I          & High\_nifty50\_weekly                                                                                              & \begin{tabular}[c]{@{}l@{}}Highest value of nifty 50 index during the \\ week previous to the Close Date\end{tabular}                             \\ \hline
B          & I          & High\_vix\_daily                                                                                                   & \begin{tabular}[c]{@{}l@{}}Highest value of vix index during the day \\ previous to the Close Date\end{tabular}                                   \\ \hline
B          & I          & High\_vix\_monthly                                                                                                 & \begin{tabular}[c]{@{}l@{}}Highest value of vix index during the \\ month previous to the Close Date\end{tabular}                                 \\ \hline
B          & I          & High\_vix\_weekly                                                                                                  & \begin{tabular}[c]{@{}l@{}}Highest value of vix index during the \\ week previous to the Close Date\end{tabular}                                  \\ \hline
B          & I          & Low\_nifty50\_daily                                                                                                & \begin{tabular}[c]{@{}l@{}}Lowest Price of nifty 50 index during \\ the day previous to the Close Date\end{tabular}                               \\ \hline
B          & I          & Low\_nifty50\_monthly                                                                                              & \begin{tabular}[c]{@{}l@{}}Lowest Price of nifty 50 index during \\ the month previous to the Close Date\end{tabular}                             \\ \hline
B          & I          & Low\_nifty50\_weekly                                                                                               & \begin{tabular}[c]{@{}l@{}}Lowest Price of nifty 50 index during \\ the week previous to the Close Date\end{tabular}                              \\ \hline
B          & I          & Low\_vix\_daily                                                                                                    & \begin{tabular}[c]{@{}l@{}}Lowest Price of vix during the day \\ previous to the Close Date\end{tabular}                                          \\ \hline
B          & I          & Low\_vix\_monthly                                                                                                  & \begin{tabular}[c]{@{}l@{}}Lowest Price of vix during the month \\ previous to the Close Date\end{tabular}                                        \\ \hline
B          & I          & Low\_vix\_weekly                                                                                                   & \begin{tabular}[c]{@{}l@{}}Lowest Price of vix during the week \\ previous to the Close Date\end{tabular}                                         \\ \hline
B          & I          & Price\_nifty50\_daily                                                                                              & \begin{tabular}[c]{@{}l@{}}Closing Price of nifty 50 index during the \\ day previous to the Close Date\end{tabular}                              \\ \hline
B          & I          & Price\_nifty50\_monthly                                                                                            & \begin{tabular}[c]{@{}l@{}}Closing Price of nifty 50 index during the \\ month previous to the Close Date\end{tabular}                            \\ \hline
B          & I          & Price\_nifty50\_weekly                                                                                             & \begin{tabular}[c]{@{}l@{}}Closing Price of nifty 50 index during the \\ week previous to the Close Date\end{tabular}                             \\ \hline
B          & I          & Price\_vix\_daily                                                                                                  & \begin{tabular}[c]{@{}l@{}}Closing Price of vix during the day \\ previous to the Close Date\end{tabular}                                         \\ \hline
B          & I          & Price\_vix\_monthly                                                                                                & \begin{tabular}[c]{@{}l@{}}Closing Price of vix during the month \\ previous to the Close Date\end{tabular}                                       \\ \hline
B          & I          & Price\_vix\_weekly                                                                                                 & \begin{tabular}[c]{@{}l@{}}Closing Price of vix during the week \\ previous to the Close Date\end{tabular}                                        \\ \hline
B          & I          & Volume\_nifty50\_daily                                                                                             & \begin{tabular}[c]{@{}l@{}}Volume traded in nifty 50 during the day \\ previous to the Close Date\end{tabular}                                    \\ \hline
B          & I          & Volume\_nifty50\_monthly                                                                                           & \begin{tabular}[c]{@{}l@{}}Volume traded in nifty 50 during the month \\ previous to the Close Date\end{tabular}                                  \\ \hline
B          & I          & Volume\_nifty50\_weekly                                                                                            & \begin{tabular}[c]{@{}l@{}}Volume traded in nifty 50 during the week \\ previous to the Close Date\end{tabular}                                   \\ \hline
B          & I          & dynamic\_last\_90Day\_success\_close                                                                               & \begin{tabular}[c]{@{}l@{}}Average success rate calculated using close price \\ in the last 90 days prior to IPO close day\end{tabular}           \\ \hline
B          & I          & dynamic\_last\_90Day\_success\_high                                                                                & \begin{tabular}[c]{@{}l@{}}Average success rate calculated using high price\\ in the last 90 days prior to IPO close day\end{tabular}             \\ \hline
B          & I          & dynamic\_last\_90Day\_success\_open                                                                                & \begin{tabular}[c]{@{}l@{}}Average success rate calculated using open price\\ in the last 90 days prior to IPO close day\end{tabular}             \\ \hline
B          & I          & previous\_quarter                                                                                                  & Quarter before the Close Date                                                                                                                     \\ \hline
B          & I          & previous\_quarter\_success\_close                                                                                  & \begin{tabular}[c]{@{}l@{}}Average success rate calculated using close price\\ in the previous calendar quarter\end{tabular}                      \\ \hline
B          & I          & previous\_quarter\_success\_high                                                                                   & \begin{tabular}[c]{@{}l@{}}Average success rate calculated using high price\\ in the previous calendar quarter\end{tabular}                       \\ \hline
B          & I          & previous\_quarter\_success\_open                                                                                   & \begin{tabular}[c]{@{}l@{}}Average success rate calculated using open price \\ in the previous calendar quarter\end{tabular}                      \\ \hline
B          & I          & \begin{tabular}[c]{@{}l@{}}Foreign direct investment, \\ net (BoP, current US\$)\end{tabular}                      & \begin{tabular}[c]{@{}l@{}}Net Foreign Direct Investment happened \\ in the year prior to IPO\end{tabular}                                         \\ \hline
B          & I          & \begin{tabular}[c]{@{}l@{}}Foreign direct investment,\\ net inflows (BoP, current US\$)\end{tabular}               & \begin{tabular}[c]{@{}l@{}}Net inflow from Foreign Direct Investment \\ happened in the year prior to IPO\end{tabular}                             \\ \hline
B          & I          & GDP (current US\$)                                                                                                 & GDP of India in the year prior to IPO                                                                                                             \\ \hline
B          & I          & GDP growth (annual \%)                                                                                             & \begin{tabular}[c]{@{}l@{}}Annual growth \% in GDP of India in the \\ year prior to IPO\end{tabular}                                              \\ \hline
B          & I          & GDP per capita growth (annual \%)                                                                                  & \begin{tabular}[c]{@{}l@{}}GDP per capita growth as annual \% in the \\ year prior to IPO\end{tabular}                                            \\ \hline
B          & I          & GNI (current US\$)                                                                                                 & Gross National Income in the year prior to IPO                                                                                                    \\ \hline
B          & I          & GNI per capita growth (annual \%)                                                                                  & \begin{tabular}[c]{@{}l@{}}Gross National Income per capita growth as \\ annual \% in the year prior to IPO\end{tabular}                           \\ \hline
B          & I          & Inflation, consumer prices (annual \%)                                                                             & Inflation rate in the year prior to the IPO                                                                                                       \\ \hline
B          & I          & Personal remittances, received (\% of GDP)                                                                         & \begin{tabular}[c]{@{}l@{}}Received personal remittances as \% of GDP \\ on the year prior to IPO\end{tabular}                                   \\ \hline
B          & I          & Initiation of Refunds                                                                                              & \begin{tabular}[c]{@{}l@{}}Date on which refunds are initiated to the \\ investors who are not allocated any shares\end{tabular}                  \\ \hline
B          & I          & \begin{tabular}[c]{@{}l@{}}Net trade in goods and \\ services (BoP, current US\$)\end{tabular}                     & \begin{tabular}[c]{@{}l@{}}Net trade in goods and services in the \\ year prior to the IPO\end{tabular}                                           \\ \hline
B          & I          & \begin{tabular}[c]{@{}l@{}}Unemployment, \\ total (\% of total labor force) \\ (modelled ILO estimate)\end{tabular} & Unemployment rate in the year prior to the IPO                                                                                                    \\ \hline
B & I & answer\_of\_question\_\textless{}n\textgreater{} & Answer generated using LLM for the n\textsuperscript{th} question. 1$<=$n$<=$25 \\ \hline
S          & D          & BSE\_High                                                                                                          & Highest price of the stock on BSE on the Listing Day                                                                                              \\ \hline
S          & D          & BSE\_Low                                                                                                           & Lowest price of the stock on BSE on the Listing Day                                                                                               \\ \hline
S          & D          & BSE\_Open                                                                                                          & Opening price of the stock on BSE on the Listing Day                                                                                              \\ \hline
B          & D          & BSE\_Last\_Trade                                                                                                   & Closing price on BSE of the IPO on the Listing Day                                                                                                \\ \hline
B          & D          & NSE\_High                                                                                                          & Highest price of the stock on NSE on the Listing Day                                                                                              \\ \hline
B          & D          & NSE\_Last\_Trade                                                                                                   & Closing price of the stock on NSE on the Listing Day                                                                                              \\ \hline
B          & D          & NSE\_Low                                                                                                           & Lowest price of the stock on NSE on the Listing Day                                                                                               \\ \hline
B          & D          & NSE\_Open                                                                                                          & Opening price of the stock on NSE on the Listing Day                                                                                              \\ \hline
S          & D          & High                                                                                                               & Overall highest price of the SME IPO on Listing day                                                                                               \\ \hline
S          & D          & Last\_Trade                                                                                                        & Overall closing price of the SME IPO on Listing Day                                                                                               \\ \hline
S          & D          & Low                                                                                                                & Overall lowest price of the SME IPO on Listing Day                                                                                                \\ \hline
S          & D          & Open                                                                                                               & Overall open price of the SME IPO on listing day                                                                                                  \\ \hline
B          & D          & Success\_Close                                                                                                     & \begin{tabular}[c]{@{}l@{}}1 if closing price on listing day is more\\ than issue price else 0\end{tabular}                                        \\ \hline
B          & D          & Success\_High                                                                                                      & \begin{tabular}[c]{@{}l@{}}1 if highest price on listing day is more\\ than issue price else 0\end{tabular}                                        \\ \hline
B          & D          & Success\_Open                                                                                                      & \begin{tabular}[c]{@{}l@{}}1 if opening price on listing day is more\\ than issue price else 0\end{tabular}                                        \\ \hline
M          & D          & elasticity  [Not used in this study]                                                                                                       & ((open price - issue price)/issue price)/(subscription rate - 1)                                                                                  \\ \hline
M          & D          & Total\_subscriptions                                                                                               & Total subscriptions the IPO received                                                                                                              \\ \hline
\end{longtable}
\end{landscape}

\section{Prompts}
\label{appendix-prompts}
\subsection{Prompt for generating answers from prospectus}
The prompt we used for generating answer using Llama-3 3b is as follows:\\
``\textit{You are an expert financial analyst who have extensive experience of participating in Initial Public Offerings (IPOs) of Indian companies. Relevant contents from Red Herring Prospectus (RHP) of an Indian company going for IPO is given to you. Your task is to analyse and answer the given question in less than 300 words as free text. Use just the content provided to you to answer the question and not anything else. If the contents are not relevant, just return the word `None'. \\ CONTENT-1: \{semantic-content\} \\ CONTENT-2: \{syntactic-content\} \\ Question: \{question\}}''

Here, \{semantic-content\} refers to the relevant information extracted using cosine similarity and \{syntactic-content\} refers to the relevant information retrieved using BM25 algorithm \cite{bm25s}.

\subsection{Prompt for estimating success of IPOs and under pricing}
We used the same prompt for Gemini 1.5 flash \cite{reid2024gemini} and Llama 3.2 3b to predict success of IPOs and under pricing. The prompt is as follows:\\
\textbf{For predicting Success}\\
``\textit{You are an expert financial analyst who have extensive experience of participating in Initial Public Offerings (IPOs) of Indian companies. You are given various facts of a company in JSON format where each key represents the type of content and value content itself. Your task is to analyse these content and predict if the (Open/Close/Highest) price of the IPO on the listing day will be more than the Issue price. Answer 1 if if the (Open/Close/Highest) price of the IPO on the listing day will be more than the Issue price, otherwise answer 0. If you are not confident answer -1. Your answer should be in -1, 0, 1 only. \\ JSON CONTENT: \{json\_content\} \\ Descriptions of keys of the JSON CONTENT are: \{col\_desc\_dict\} \\ Response:}''

\textbf{For predicting under-pricing}
``\textit{You are an expert financial analyst who have extensive experience of participating in Initial Public Offerings (IPOs) of Indian companies. You are given various facts of a company in JSON format where each key represents the type of content and value content itself. Your task is to analyse these content and predict if the under-pricing with respect to (Open or Close or Highest) price of the IPO on the listing day i.e. (Open or Close or Highest - issue price)/(issue price). Answer should be a real number only. If you are not confident answer nan. \\ JSON CONTENT: \{json\_content\} \\ Descriptions of keys of the JSON CONTENT are: \{col\_desc\_dict\} \\ Response:}''

\end{document}